\pdfoutput=1
\documentclass[11pt,a4paper]{article}
\usepackage[hyperref]{eacl2021}
\usepackage{times}
\usepackage{latexsym}

\usepackage{graphicx, subcaption}
\usepackage{amsmath}
\usepackage{booktabs, tabularx}
\usepackage{multirow}
\usepackage{amsfonts}
\usepackage{tablefootnote}
\usepackage{microtype}
\usepackage{subcaption}
\aclfinalcopy % Uncomment this line for the final submission
\title{SpanEmo: Casting Multi-label Emotion Classification as Span-prediction}

\author{Hassan Alhuzali \quad \quad Sophia Ananiadou \\
    National Centre for Text Mining \\
    Department of Computer Science, The University of Manchester, United Kingdom \\
  \texttt{\{hassan.alhuzali@postgrad.,sophia.ananiadou@\}manchester.ac.uk}}

\begin{document}
\raggedbottom

%\setcode{utf8}
%\setarab
%\novocalize

\maketitle
\begin{abstract}
  Emotion recognition (\textsc{er}) is an important task in Natural Language Processing (\textsc{nlp}), due to its high impact in real-world applications from health and well-being to author profiling, consumer analysis and security.~Current approaches to \textsc{er}, mainly classify emotions independently without considering that emotions can co-exist. Such approaches overlook potential ambiguities, in which multiple emotions overlap.~We propose a new model ``SpanEmo'' casting multi-label emotion classification as span-prediction, which can aid \textsc{er} models to learn associations between labels and words in a sentence.~Furthermore, we introduce a loss function focused on modelling multiple co-existing emotions in the input sentence. Experiments performed on the SemEval2018 multi-label emotion data over three language sets (i.e., English, Arabic and Spanish) demonstrate our method's effectiveness.~Finally, we present different analyses that illustrate the benefits of our method in terms of improving the model performance and learning meaningful associations between emotion classes and words in the sentence\footnote{Source code is available at \url{https://github.com/hasanhuz/SpanEmo}}.  %, extracting representative words per emotion
\end{abstract}

\section{Introduction}
Emotion is essential to human communication, thus emotion recognition (\textsc{er}) models have a host of applications from health and well-being~\cite{alhuzali-ananiadou-2019-improving,aragon2019detecting,chen2018mood} to consumer analysis~\cite{alaluf2019emotions,herzig2016predicting} and user profiling~\cite{volkova2016inferring,mohammad2013using}, among others. Interest in this area has given rise to new \textsc{nlp} approaches aimed at emotion classification, including single-label and multi-label emotion classification. Most existing approaches for multi-label emotion classification~\cite{ying-etal-2019-improving,baziotis2018ntua,yu2018improving,badaro-etal-2018-ema,mulki-etal-2018-tw,Mohammad2018semeval,yang2018sgm} do not effectively capture emotion-specific associations, which can be useful for prediction, as well as learning of association between emotion labels and words in a sentence. In addition, standard approaches in emotion classification treat individual emotion independently. However, emotions are not independent; a specific emotive expression can be associated with multiple emotions. The existence of association/correlation among emotions has been well-studied in psychological theories of emotions, such as Plutchik's wheels of emotion~\cite{plutchik1984emotions} that introduces the notion of mixed and contrastive emotions. For example, ``joy'' is close to ``love'' and ``optimism'', instead of ``anger'' and ``sadness''. %Taking emotion correlations into consideration can improve model performance as well as address the ambiguity characteristic of the task, especially for those correlated emotions.

%the circumplex model defines emotion on a two-dimensional space of valance (positiveness-negativeness) and arousal (active-passive), highlighting that emotions within the same valance (e.g. anger, fear and sad) tend to occur together more often than emotions from the different valance.~Incorporating emotion occurrences can address the ambiguity characteristic of the task, especially for those correlated emotions.%Taking into account such co-occurrences between emotions is crucial for multi-label emotion classification in addressing the ambiguity characteristic of the task, especially for those highly correlated emotions.%This supports our notion that emotions are indeed not independent. and Russell's circumplex model~\cite{russell1980circumplex}

%Thus's wee, incorporating label dependenciePlutchiks is crucial so as to maximise the distance between positive and negative labels. Table~\ref{intro} shows a couple of examples.\footnote{We specifically mean to learn/capture associations between emotion labels and words in a sentence.}
% Table generated by Excel2LaTeX from sheet 'Sheet6' \textsc{woe} designs eight sectors to indicate that there are eight primary emotion dimensions, each of which has a polar opposite as well as an intensity degree. It also introduces that an emotion (e.g. submission and love) can be a mix of two primary emotions. In a similar vein, 
\begin{table}[htbp]
  \centering
  \scalebox{0.9}{
    \begin{tabular}{cp{12em}p{4em}}
    \toprule
    \textbf{\#} & \textbf{Sentence} & \textbf{GT} \\
    \midrule
    S1 &  {well my day started off great} the mocha machine wasn't working @ mcdonalds. & anger, disgust, joy, sadness \\
    \midrule
    S2 & I'm doing all this to make sure you smiling down on me bro. & joy, love, optimism \\
    \bottomrule
    \end{tabular}}
    \caption{Example Tweets from SemEval-18 Task 1. GT represents the ground truth labels.}
  \label{intro}%
\end{table}%

Consider \textsc{s}1 in Table~\ref{intro}, which contains a mix of positive and negative emotions, although it is more negative oriented. This can be observed clearly via the ground truth labels assigned to this example, where the first part of this sentence only expresses a positive emotion (i.e., joy), while the other part expresses negative emotions.~For example, clue words like ``great'' are more likely to be associated with ``joy'', whereas ``wasn't working'' are more likely to be associated with negative emotions. Learning such associations between emotion labels and words in a sentence can help \textsc{er} models to predict the correct labels. \textsc{s}2 further highlights that certain emotions are more likely to be associated with each other.~\newcite{MohammadB17starsem} also observed that negative emotions are highly associated with each other, while less associated with positive emotions. Based on these observations, we seek to answer the following research questions: i) how to enable ER models to learn emotion-specific associations by taking into account label information and ii) how to benefit from the multiple co-existing emotions in a multi-label emotion data set with the intention of learning label correlations. Our contributions are summarised as follows:

%Consider \textsc{s}1 in Table~\ref{intro}, which shares a mix of positive and negative emotions although it is more negative oriented. This can be clearly observed via the assigned ground truth labels to this example, where the first part of this sentence only expresses a positive emotion (i.e., joy), while the other part expresses negative emotions. \textsc{s}2 further highlights that certain emotions are more likely to be associated with each other.~\newcite{MohammadB17starsem} also observes that negative emotions are highly associated with each other, while less associated with positive emotions. Based on these observations, we seek to answer the following research questions: i) how to learn emotion-specific associations by taking into account label information and ii) how to benefit from the multiple co-existing emotions in a multi-label emotion data set with the intention of learning label dependencies. Our contributions are summarised as follows:

\noindent
\textbf{I.} a novel framework casting the task of multi-label emotion classification as a span-prediction problem.~We introduce ``\textsc{s}pan\textsc{e}mo'' to train the model to take into consideration both the input sentence and a label set (i.e., emotion classes) for selecting a span of emotion classes in the label set as the output. The objective of \textsc{s}pan\textsc{e}mo is to predict emotion classes directly from the label set and capture associations corresponding to each emotion. %In this respect, treat multi-label classification as span-prediction  multiple tokens from the label set, similar to questions-answering task in transformers-based models. are defined in  labels and an input sentence into the BERT encoder as to enable it to attend to both information at the same time, hoping this will lead to improving the input sentence representation with respect to each emotion label, and hence enable the encoder to capture the context information related to the corresponding emotion label. to predict a span of text (i.e., emotion classes) in the label set as the output for multi-label emotion classification.

% The task is formalised in this way because \textsc{s}pan\textsc{e}mo selects its final predictions directly from the label set included as part of the input in our framework.~In this respect, 

% incorporate a label set (i.e., emotion classes) into the input sentence, which will be eventually used to predict multiple labels directly from the label set.
    
\noindent
\textbf{II.} a loss function, modelling multiple co-existing emotions for each input sentence.~We make use of the label-correlation aware loss (\textsc{lca})~\cite{yeh2017learning}, originally introduced by~\newcite{zhang2006multilabel}. The objective of this loss function is to maximise the distance between positive and negative labels, which is learned directly from the multi-label emotion data set. %We demonstrate its effectiveness in learning emotion correlations, which in turns improve the model performance.   %The objective of this loss function is to maximise the distance between positive and negative labels, which is learned directly from the multi-label emotion data set.

\noindent
\textbf{III.} a large number of experiments and analyses both at the word- and sentence-level, demonstrating the strength of \textsc{s}pan\textsc{e}mo for multi-label emotion classification across three languages (i.e. English, Arabic and Spanish).

% Empirical evaluation demonstrates that our method can enhance model performance compared to previous approaches of multi-label emotion classification, even without relying on any external resources. Adding label representation to the input-sentence helps the model learn meaningful associations between emotion classes and capture representative words per emotion.

The rest of the paper is organised as follows: section~\ref{method} describes our methodology, while section~\ref{exp} discusses experimental details. We evaluate the proposed method and compare it to related methods in section~\ref{results_sec}. Section~\ref{ana_sec} reports on the analysis of results, while section~\ref{RW} reviews related work. We conclude in section~\ref{conc}.

\section{Methodology} \label{method}
%the open-source Hugging Face implementation~\cite{Wolf2019HuggingFacesTS}
%This paragraph needs rephrasing?????
%The overall architecture of our span-based multitask framework (SpanMlt) is shown in Figure 2.Given an input sentence, a base encoder is adopted to learn contextualized word representations. Then, a span generator is deployed to enumerate all possible spans, which are represented based on the hidden outputs of the base encoder. For the multitask learning setup, the span representations are shared for two output scorers. The term scorer is to assign the term label with the highest score to each span. And the relation scorer is to evaluate the pair-wise correspondence between every two spans and assign a binary label to each span pair.
\subsection{Framework}
%Next, a feed forward network (FFN) is used to project the learned representations into a single score for each token. We can then use the scores for the label tokens as predictions for the corresponding label, since there is a 1-to-1 correspondence between the label tokens and the original emotion labels.

\begin{figure}[ht]
\centering
   \includegraphics[width=0.95\linewidth, trim=0cm .3cm 0cm .2cm, clip]{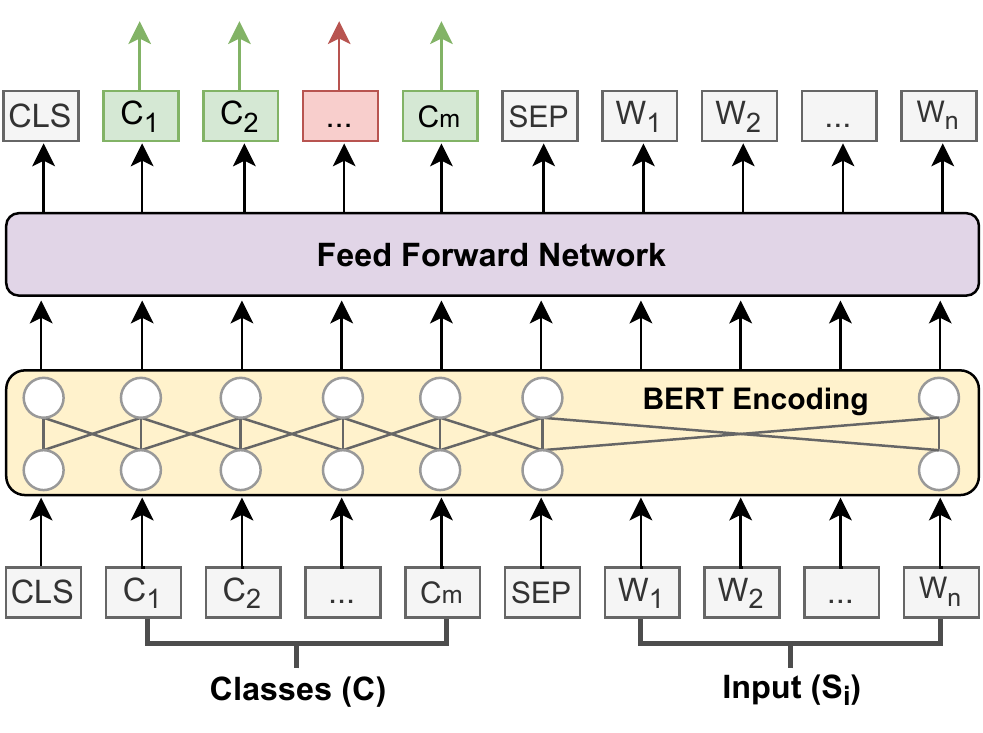}
  \caption{Illustration of our proposed framework (SpanEmo).}\label{arch}
\end{figure}

Figure~\ref{arch} presents our framework (\textsc{s}pan\textsc{e}mo). Given an input sentence and a set of classes, a base encoder was employed to learn contextualised word representations. Next, a feed forward network (\textsc{ffn}) was used to project the learned representations into a single score for each token. We then used the scores for the label tokens as predictions for the corresponding emotion label. The green boxes at the top of the \textsc{ffn} illustrate the positive label set, while the red ones illustrate the negative label set for multi-label emotion classification. We now turn to describing our framework in detail. 

%, which is a pretrained deep bidirectional model that obtains competitive results across various tasks as the encoder.  Our framework receives two segments (i.e., classes and an input sentence). The objective of BERT encoder is first to interpolate between the two segments and then to learn hidden representation for each token. Next, the output is passed to FFN, which transforms the representation into a single score for each token. Finally, the classes segment is used for predictions, where green boxes illustrate the positive label set, while red boxes illustrate the negative label set for multi-label emotion classification.

\subsection{Our Method (\textsc{s}pan\textsc{e}mo)}
%\{c_{1} \dots c_{m}\}
%\{w_{1} \dots w_{n}\}
Let $\{(s_{i}, y_{i})\}_{i=1}^{N}$ be a set of \textsc{n} examples with the corresponding emotion labels of \textsc{c} classes, where $s_{i}$ denotes the input sentence and $y_{i} \in \{0,1\}^{m}$ represents the label set for $s_{i}$. As shown in Figure~\ref{arch}, both the label set and the input sentence were passed into the encoder \textsc{bert}~\cite{devlin2019bert}. The encoder received two segments: the first corresponds to the set of emotion classes, while the second refers to the input sentence. The hidden representations ($H_{i}~\in~R^{T~\times~D}$)\footnote{$\textsc{t}$ and $\textsc{d}$ denote the input length and dimensional size, respectively.} for each input sentence and the label set were obtained as follows:\begin{equation}
    \textsc{h}_{i} = \text{Encoder}(\mathrm{[CLS]} + \mathrm{|C|} + \mathrm{[SEP]} + \mathbf{s}_{i}),
\end{equation}
where $\{[\textsc{cls}]$, $[\textsc{sep}]\}$ are special tokens and $|\textsc{c}|$ denotes the size of emotion classes. Feeding both segments to the encoder has a few advantages. First, the encoder can interpolate between emotion classes and all words in the input sentence. Second, a hidden representation is generated both for words and emotion classes, which can be further used to understand whether the encoder can learn association between the emotion classes and words in the input sentence. Third, \textsc{s}pan\textsc{e}mo is flexible because its predictions are directly produced from the first segment corresponding to the emotion classes.%~This makes \textsc{s}pan\textsc{e}mo distinct from most multi-label emotion classification, depending only on the $[\textsc{cls}]$ token representation for making its final predictions.

We further introduced a feed-forward network (\textsc{ffn}) consisting of a non-linear hidden layer with a \textsc{t}anh activation ($f_{i}(\textsc{h}_{i})$) as well as a position vector $p_{i} \in R^{D}$, which was used to compute a dot product between the output of $f_{i}$ and $p_{i}$. As our task involved a multi-label emotion classification, we added a sigmoid activation to determine whether $\text{class}_{i}$ was the correct emotion label or not. It should be mentioned that the use of the position vector is quite similar to how start and end vectors are defined in transformer-based models for question-answering. Finally, the span-prediction tokens were obtained from the label segment and then compared with the ground truth labels since there was a 1-to-1 correspondence between the label tokens and the original emotion labels.
\begin{equation}
    \mathbf{\hat{y}} = \mathrm{sigmoid}(\text{FFN}(\mathbf{H}_{i})),
\end{equation}

\subsection{Label-Correlation Aware (LCA) Loss}
Following~\newcite{yeh2017learning}, we employed the label-correlation aware loss, which takes a vector of true-binary labels ($y$), as well as a vector of probabilities ($\hat y$), as input:
\begin{equation}
\small
\mathcal{L}_\text{LCA}(\mathbf{y}, \hat{\mathbf{y}})=\frac{1}{\left|\mathbf{y}^{0}\right|\left|\mathbf{y}^{1}\right|} \sum_{(p, q) \in \mathbf{y}^{0} \times \mathbf{y}^{1}} \exp \left(\hat{\mathbf{y}}_{p}-\hat{\mathbf{y}}_{q}\right),
\end{equation}
where $y^{0}$ denotes the set of negative labels, while $y^{1}$ denotes the set of positive labels. $\hat y_{p}$ represents the $p^{th}$ element of vector $\hat y$. The objective of this loss function is to maximise the distance between positive and negative labels by implicitly retaining the label-dependency information. In other words, the model should be penalised when it predicts a pair of labels that should not co-exist for a given example.

\subsection{Training Objective}
% we experimentally observed that training “spanemo” jointly helped the model to achieve the best results
%Since the objective of LCA is to maximise the distance between positive and negative label set, we combine this loss function with binary cross-entropy (\textsc{bce}) so as to maximises the probability of correct labels. We combine LCA loss with binary crossentropy loss (BCE) and train them jointly as follows
To model label-correlation, we combined \textsc{lca} loss with binary cross-entropy (\textsc{bce}) and trained them jointly. This aimed to help the \textsc{lca} loss to focus on maximising the distance between positive and negative label sets, while at the same time taking advantage of the \textsc{bce} loss through maximising the probability of the correct labels. We experimentally observed that training our approach jointly with those two loss functions produced the best results. The overall training objective was computed as follows: 
\begin{equation}\label{equ:totloss}
\mathcal{L}= 
(1 - \alpha) \mathcal{L}_\text{BCE} +
\alpha \sum_{i=1}^{M} \mathcal{L}_\text{LCA},
\end{equation}
where $\alpha \in [0,1]$ denotes the weight used to control the contribution of each part to the overall loss.
% (\mathbf{Y}, \hat{\mathbf{Y}})

\section{Experiments} \label{exp}

\subsection{Implementation Details} \label{imp_det}
We used \textsc{p}y\textsc{t}orch~\cite{paszke2017automatic} for implementation and ran all experiments on an \textsc{n}vidia \textsc{g}e\textsc{f}orce \textsc{gtx} 1080 with 11 \textsc{gb} memory. We also trained \textsc{bert$_{\text{base}}$} utilising the open-source Hugging-Face implementation~\cite{Wolf2019HuggingFacesTS}. For experiments related to \textsc{a}rabic, we chose ``bert-base-arabic'' developed by~\newcite{safaya2020kuisail}, while selecting ``bert-base-spanish-uncased'' developed by~\newcite{CaneteCFP2020} for \textsc{s}panish. All three models were trained on the same hyper-parameters with a fixed initialisation seed, including a feature dimension of $786$, a batch size of $32$, a dropout rate of $0.1$, an early stop patience of $10$ and $20$ epochs. Adam was selected for optimisation~\cite{kingma2014adam} with a learning rate of $2e\text{-}5$ for the BERT encoder, and a learning rate of $1e\text{-}3$ for the FFN. It should be mentioned that we tuned our method only on the validation set and further report on the analysis of the effect of parameter $\alpha$ in section~\ref{alpha_sec}. Table~\ref{hyper} summarises the hyper-parameters used in our experiments.  

% \iffalse
\begin{table}[h]
\centering 
\scalebox{0.95}{
\begin{tabular}{l|c}\toprule
\textbf{Parameter} & \textbf{Value}\\ \midrule
{Feature dimension}  & $768$\\
Batch size & $32$  \\
Dropout   & $0.1$\\
Early stop patience & $10$ \\
Number of epochs & $20$ \\
lr-BERT & $2e\text{-}5$ \\
lr-FFN & $1e\text{-}3$ \\
Optimiser & Adam \\
Alpha ($\alpha$) & $0.2$ \\
\bottomrule
\end{tabular}}
\caption{Hyper-parameter values.~lr: refers to the Learning rate.}\label{hyper}
\end{table}
% \fi

\subsection{Data Set and Task Settings}\label{data}
In this work, we chose \textssc{s}em\textsc{e}val2018~\cite{Mohammad2018semeval} for our multi-label emotion classification, which is based on labelled data from tweets in \textsc{e}nglish, \textsc{a}rabic and \textsc{s}panish.~The data was originally partitioned into three sets: training set (\textsc{t}rain), validation set (\textsc{v}alid) and test (\textsc{t}est) set. Following the metrics in~\newcite{Mohammad2018semeval}, we run our experiments on micro F1-score, macro F1-score and \textsc{j}accard index score\footnote{jacS is defined as the size of the intersection divided by the size of the union of the true label set and predicted label set.}. Table~\ref{data stats} presents the summary of all three sets for each language, including the number of instances in the train, valid and test sets. In addition, the number of emotion classes and the percentage of instances with varying numbers of classes (co-existing) are included. It is worth noting that these percentages do not include the neutral instances.% the percentages of co-existing emotions do not sum up to $100$ because they do not contain neutral instances.
% Table generated by Excel2LaTeX from sheet 'data statistics'
\begin{table}[htbp]
  \centering
  \scalebox{0.9}{
    \begin{tabular}{l|ccc}
    \toprule
    \textbf{Info./Lang.} & \textbf{English} & \textbf{Arabic} & \textbf{Spanish} \\ \midrule
    {Train (\#)} & 6,838 & 2,278 & 3,561 \\
    {Valid (\#)}  & 886   & 585   & 679 \\
    {Test (\#)} & 3,259 & 1,518 & 2,854 \\
    {Total (\#)} & 10,983 & 4,381 & 7,094 \\ 
    {Classes} (\#)  & 11 & 11 & 11 \\\bottomrule
    {$1$ co.emo (\%)} & 14.36 & 21.38 & 39.11\\
    {$2$ co.emo (\%)} & 40.55 & 39.03 & 42.15\\
    {$3$ co.emo (\%)} & 30.92 & 29.85& 12.76\\
    \bottomrule
    \end{tabular}}
    \caption{Data Statistics. co.emo: refers to the percentage of co-existing emotions.}
  \label{data stats}%
\end{table}%

To pre-process the data, we utilised a tool designed for the specific characteristics of \textsc{t}witter, i.e., misspellings and abbreviations~\cite{baziotis2017datastories}.~The tool offers different functionalities, such as tokenisation, normalisation, spelling correction, and segmentation.~We used the tool to tokenise the text, convert words to lower case, normalise user mentions, urls and repeated-characters.

\subsection{Multi-label Emotion Classification}\label{bl}
We compared the performance of~\textsc{s}pan\textsc{e}mo to some baseline as well as state-of-the-art models on all three languages. For experiments related to \textsc{e}nglish, we selected seven models, while we chose three models for both \textsc{a}rabic and \textsc{s}panish.~We also include the results of \textsc{bert$_{\text{base}}$}.

\subsubsection{English}
\textsc{e}nglish models include
\textsc{jbnn}~\cite{he2018joint}, \textsc{datn}~\cite{yu2018improving}, \textsc{ntua}~\cite{baziotis2018ntua}, \textsc{rer}c~\cite{zhou2018relevant}, \textsc{bert$_{\text{base}}$}+\textsc{dk}~\cite{ying-etal-2019-improving}, \textsc{bert$_{\text{base}}$-gcn}~\cite{xu2020emograph} and \textsc{lem}~\cite{fei2020latent}. \textsc{jbnn} introduces a joint binary neural network, which focuses on learning the relations between emotions based on the theory of \textsc{p}lutchik’s wheel of emotions~\cite{plutchik1980emotion}, and then performing multi-label emotion classification via integrating these label relations into the loss function.~\textsc{datn} proposes a dual attention transfer network to improve multi-label emotion classification with the help of sentiment classification, while \textsc{ntua} is ranked the top-1 model of the SemEval2018 competition as it relies on different pre-training and fine-tuning strategies. \textsc{rer}c defines a ranking emotion relevant loss focused on incorporating emotion relations into the loss function to improve both emotion prediction and rankings of relevant emotions.~Both \textsc{bert$_{\text{base}}$}+\textsc{dk} and \textsc{bert$_{\text{base}}$-gcn} utilise the same encoder as our own with the former considering additional domain knowledge (\textsc{dk}) and the latter capturing emotion relations through Graph Convolutional Network (\textsc{gcn}), respectively. \textsc{lem} introduces a latent emotion memory network, in which the latent emotion module learns emotion distribution via a variational autoencoder, while the memory module captures features corresponding to each emotion.

% Both \textsc{bert$_{\text{base}}$}+\textsc{dk} and \textsc{bert-gcn} utilise the same encoder as our own, but the former considers additional domain knowledge (\textsc{dk}), while the latter captures emotion relations through graph convolutional network (\textsc{gcn}).
%EDL: The third model introduces emotion distribution learning (\textsc{edl}), which focused on learning the relations between emotions based on the theory of \textsc{p}lutchik’s wheel of emotions~\cite{plutchik1980emotion}, and then performed multi-label emotion classification via integrating these label relations into the loss function.
\subsubsection{Arabic}
Arabic models consist of \textsc{ema}~\cite{badaro-etal-2018-ema}, \textsc{t}w-\textsc{s}t\textsc{ar}~\cite{mulki-etal-2018-tw} and \textsc{hef}~\cite{alswaidan2020hybrid}.~\textsc{ema} is the best performing model from the \textsc{s}em\textsc{e}val2018 competition on this set. It utilises various pre-processing steps (e.g. diacritics removal, normalisation, emojis transcription and stemming), as well as different classification algorithms. The~\textsc{t}w-\textsc{s}t\textsc{ar} model applies some pre-processing steps and then uses \textsc{tf-idf} to learn features of a \textsc{s}upport \textsc{v}ector \textsc{m}achine. \textsc{hef} is based on a hybrid neural network, including different word embeddings (e.g. \textsc{w}ord2\textsc{v}ec, \textsc{g}love, \textsc{f}ast\textsc{t}ext) plus variations of \textsc{rnn} neural networks, such as \textsc{l}ong \textsc{s}hort-\textsc{t}erm \textsc{m}emory and \textsc{g}ated \textsc{r}ecurrent \textsc{u}nit.

\subsubsection{Spanish}
\textsc{s}panish models comprise \textsc{t}w-\textsc{s}t\textsc{ar}~\cite{mulki-etal-2018-tw}, \textsc{el}i\textsc{rf}~\cite{gonzalez-etal-2018-elirf} and \textsc{milab}~\cite{Mohammad2018semeval}.~The~\textsc{el}i\textsc{rf} model applies some pre-processing steps while also adapting the tweet tokeniser for \textsc{s}panish tweets. \textsc{milab} is the best performing model from the \textsc{s}em\textsc{e}val2018 shared-task on this set.  

%\textsc{lem} model obtains the best macro f1-score among the three compared models on the \textsc{e}nglish set, while the \textsc{ntua-slp} model achieves the highest micro f1-score and accuracy. \textsc{hef+df}, which is developed for Arabic emotion data, shows the best performance among the three baseline models on all metrics. For experiments ran on Spanish from tweets, \textsc{milab-snu} achieves higher micro f1-score and accuracy over the two baseline models, while \textsc{el}i\textsc{rf-upv} obtains the highest macro f1-score. However, 

\section{Results}\label{results_sec}
Table~\ref{res_} presents the performance of our proposed approach (\textsc{s}pan\textsc{e}mo) on all three languages, in terms of micro F1-score (miF1), macro F1-score (maF1) and \textsc{j}accard index score (jacS), and compares it to the baseline and state-of-the-art models discussed in section~\ref{bl}.

\begin{table}[h]
\centering
\scalebox{0.95}{
\begin{tabular}{l|ccc}
\toprule
{\textbf{Language}} & \multicolumn{3}{c}{\textbf{English}}   \\\cmidrule{2-4}
{\textbf{Model/Metric}} & \textbf{miF1} & \textbf{maF1} & \textbf{jacS} \\ \midrule 
JBNN & 0.632 &  0.528  & - \\  
RERc & 0.651 &  0.539  & - \\
DATN   & -              & 0.551          & 0.583          \\
NTUA  & 0.701          & 0.528          & 0.588          \\
BERT$_{\text{base}}$ & 0.695          & 0.520          & 0.570 \\
BERT$_{\text{base}}+$DK  & \textbf{0.713}  & 0.549 & 0.591\\ 
% acc 58.9 micro 70.7 macro 56.3
BERT$_{\text{base}}$-GCN & {0.707} & {0.563} & {0.589} \\
LEM   & 0.675          & 0.567          & -   \\ 
SpanEmo (ours)   & \textbf{0.713} & \textbf{0.578} & \textbf{0.601} \\
% \midrule
% SpanEmo (ours)    & \textbf{0.713} & 0.578 & \textbf{0.601} \\
% SpanEmo\text{-}w/o LCA & 0.712          & 0.564          & 0.590 \\
% %HL: 0.1526 F1-Macro: 0.5834 F1-Micro: 0.6978 JS: 0.5824 Time: 00:12
% SpanEmo\text{-}w/o BCE &   0.698        &    \textbf{0.583}       & 0.582 \\
% BERT$_{\text{base}}$ [CLS] & 0.695          & 0.520          & 0.570 \\
\midrule \midrule
 & \multicolumn{3}{c}{\textbf{Arabic}}  \\ \cmidrule{2-4}
 &  \textbf{miF1} & \textbf{maF1} & \textbf{jacS} \\ \midrule
%\textsc{ema}~\cite{badaro-etal-2018-ema}, \textsc{t}w-\textsc{s}t\textsc{ar}~\cite{mulki-etal-2018-tw} and \textsc{hef+df}~\cite{alswaidan2020hybrid}.
Tw-StAR  & 0.597          & 0.446          & 0.465          \\
EMA    & 0.618          & 0.461          & 0.489          \\
BERT$_{\text{base}}$ & 0.650          & 0.477          & 0.523  \\
HEF  & 0.631          & 0.502          & 0.512          \\
SpanEmo (ours) & \textbf{0.666} & \textbf{0.521} & \textbf{0.548} \\
% \midrule
% SpanEmo  (ours)  & \textbf{0.666} & 0.521 & \textbf{0.548} \\
% SpanEmo\text{-}w/o LCA & 0.654          & 0.481          & 0.534 \\
% SpanEmo\text{-}w/o BCE & 0.660   &  \textbf{0.526}  & 0.532  \\
% BERT$_{\text{base}}$ [CLS] & 0.650          & 0.477          & 0.523  \\
\midrule \midrule
 & \multicolumn{3}{c}{\textbf{Spanish}}  \\ \cmidrule{2-4}
 &  \textbf{miF1} & \textbf{maF1} & \textbf{jacS} \\ \midrule
%\textsc{s}panish baseline models comprise of \textsc{t}w-\textsc{s}t\textsc{ar}~\cite{mulki-etal-2018-tw}, \textsc{el}i\textsc{rf}-\textsc{upv}~\cite{gonzalez-etal-2018-elirf} and \textsc{milab-snu}~\cite{Mohammad2018semeval}
Tw-StAR  &    0.520       &   0.392        &    0.438       \\
ELiRF  &      0.535    &   0.440        &     0.458      \\
MILAB &      0.558     &    0.407      &  0.469        \\ 
BERT$_{\text{base}}$ & 0.596 & 0.474 &  0.487  \\
SpanEmo (ours) & \textbf{0.641}  & \textbf{0.532}  & \textbf{0.532} \\
% \midrule
% SpanEmo (ours)  & \textbf{0.641}  & 0.532  & \textbf{0.532} \\
% SpanEmo\text{-}w/o LCA & 0.629 & 0.526 & 0.507 \\
% % F1-Macro: 0.5442 F1-Micro: 0.6057 JS: 0.4985 
% SpanEmo\text{-}w/o BCE & 0.606 & \textbf{0.544} & 0.499\\
% BERT$_{\text{base}}$ [CLS] & 0.596 & 0.474 &  0.487  \\
\bottomrule    
\end{tabular}}
\caption{The results of multi-label emotion classification on SemEval-2018 test set.}% The last three rows in each group corresponds to the ablation study discussed in section~\ref{ablation}.}
\label{res_}
\end{table}

As shown in Table~\ref{res_}, our method outperformed all models on all languages, as well as on almost all metrics, with a marginal improvement of up to 1-1.3\% for \textsc{e}nglish, 1.9-3.6\% for \textsc{a}rabic and 6.3-9.2\% for \textsc{s}panish. This demonstrates the utility and advantages of \textsc{s}pan\textsc{e}mo, as well as the label-correlation aware loss for improving the performance of multi-label emotion classification in \textsc{e}nglish, \textsc{a}rabic and \textsc{s}panish.

Based on the empirical results reported in Table~\ref{res_}, the following observations can be made. First, incorporating the relations between emotions into the models tends to lead to higher performance, especially for macro F1-score. For example, both \textsc{datn} and \textsc{lem} learn emotion-related features and achieve better performance than \textsc{ntua} and \textsc{bert$_{\text{base}}$}+\textsc{dk}. Additionally, \textsc{el}i\textsc{rf} makes use of various sentiment/emotion features (i.e., learned from lexica) and it yielded the best performance among the three compared models. This corroborates our earlier hypothesis that learning emotion-specific associations is crucial for improving the performance. Although \textsc{bert$_{\text{base}}$}+\textsc{dk} adopts the same encoder as our own and adds domain knowledge, our method still performs strongly, especially for both macro F1- and jaccard score with a marginal improvement of up to 2.9\% and 1\%, respectively.~In short, capturing emotion-specific associations as well as integrating the relations between emotions into the loss function, helped \textsc{s}pan\textsc{e}mo to achieve the best results compared with all models on almost all metrics.

\subsection{Ablation Study}\label{ablation}

%  where we ablated one aspect of our framework each time train on one under different settings:~firstly the model is trained without \textsc{lca} loss, secondly it is trained without \textsc{bce} loss and thirdly it is trained without the label segment. as a normal classification task (\textsc{bert$_{\text{base}} [\textsc{cls}]$})

To understand the effect of our framework, we undertook an ablation study of the model performance under three settings: firstly, the model was trained only with \textsc{bce} loss; secondly, it was trained only with \textsc{lca} loss; and thirdly it was trained without the label segment. The third setting is equivalent to training the model as a simple multi-label classification task, by only considering the input sentence.  

\begin{table}[!h]
\centering
\scalebox{0.95}{
\begin{tabular}{l|ccc}
\toprule
{\textbf{Language}} & \multicolumn{3}{c}{\textbf{English}}   \\\cmidrule{2-4}
{\textbf{Model/Metric}} & \textbf{miF1} & \textbf{maF1} & \textbf{jacS} \\ \midrule 
SpanEmo (joint)    & \textbf{0.713} & 0.578 & \textbf{0.601} \\
\;\;\; - $\mathcal{L}$ (LCA) & 0.712          & 0.564          & 0.590 \\
%HL: 0.1526 F1-Macro: 0.5834 F1-Micro: 0.6978 JS: 0.5824 Time: 00:12
\;\;\; - $\mathcal{L}$ (BCE) &   0.698        &    \textbf{0.583}       & 0.582 \\
\;\;\; - Label Seg. & 0.695          & 0.520          & 0.570 \\
% BERT$_{\text{base}}$ [CLS] & 0.695          & 0.520          & 0.570 \\
\midrule \midrule
 & \multicolumn{3}{c}{\textbf{Arabic}}  \\ \cmidrule{2-4}
 &  \textbf{miF1} & \textbf{maF1} & \textbf{jacS} \\ 
 \midrule
% BERT$_{\text{base}}$ [CLS] & 0.650          & 0.477          & 0.523  \\
SpanEmo  (joint)  & \textbf{0.666} & 0.521 & \textbf{0.548} \\
\;\;\; - $\mathcal{L}$ (LCA) & 0.654          & 0.481          & 0.534 \\
\;\;\; - $\mathcal{L}$ (BCE) & 0.660   &  \textbf{0.526}  & 0.532  \\
\;\;\; - Label Seg. & 0.650          & 0.477          & 0.523  \\
\midrule \midrule
 & \multicolumn{3}{c}{\textbf{Spanish}}  \\ \cmidrule{2-4}
 &  \textbf{miF1} & \textbf{maF1} & \textbf{jacS} \\ 
 \midrule
% BERT$_{\text{base}}$ [CLS] & 0.596 & 0.474 &  0.487  \\
SpanEmo (joint)  & \textbf{0.641}  & 0.532  & \textbf{0.532} \\
\;\;\; - $\mathcal{L}$ (LCA) & 0.629 & 0.526 & 0.507 \\
\;\;\; - $\mathcal{L}$ (BCE) & 0.606 & \textbf{0.544} & 0.499\\
\;\;\; - Label Seg. & 0.596 & 0.474 &  0.487  \\
\bottomrule    
\end{tabular}}
\caption{Ablation experiment results. The last two rows from each group correspond to the removal of the respective loss function, whereas the last row corresponding to the removal of the label segment.}% The last three rows in each group corresponds to the ablation study discussed in section~\ref{ablation}.}
\label{sec:abldation_table}
\end{table}

% whereas we secondly Each time we excluded one aspect (i.e., \textsc{lca} loss, \textsc{bce} loss and label segment) of our framework and retrained \textsc{s}pan\textsc{e}mo without that aspect. It is worth mentioning that removing the label segment is similar to training the model as a simple classification task, by only considering the $[\textsc{cls}]$ token representation. 

%%%%%%%%%%%%%%%%%%%%%%%%%%%%%%%%%%%%%%%%%%%%%%%%%%%%%%%
% Table generated by Excel2LaTeX from sheet 'co_exis_v2'
\begin{table*}[!ht]
  \centering
\scalebox{0.95}{
    \begin{tabular}{lc|ccc|ccc|ccc}
    \toprule
    \multicolumn{2}{c|}{\textbf{Model/Metric}} & \textbf{miF1}  & \textbf{maF1}  & \textbf{jacS} & \textbf{miF1}  & \textbf{maF1}  & \textbf{jacS} & \textbf{miF1}  & \textbf{maF1}  & \textbf{jacS} \\ \cmidrule{1-11}
    \textbf{English} & $\mathcal{L}$ & \multicolumn{3}{c|}{$\geq$ 1 co.emo} & \multicolumn{3}{c|}{$\geq$ 2 co.emo} & \multicolumn{3}{c}{$\geq$ 3 co.emo} \\
    \midrule
    BERT$_{\text{base}}$ & BCE & 0.703 & 0.515 & 0.587 & 0.712 & 0.521 & 0.596 & 0.692 & 0.509 & 0.554 \\
    SpanEmo & BCE & 0.716 & 0.563 & 0.599 & 0.737 & 0.578 & 0.629 & 0.748 & 0.597 & 0.639 \\
    SpanEmo & Joint & \textbf{0.724} & \textbf{0.590} & \textbf{0.613} & \textbf{0.746} & \textbf{0.606} & \textbf{0.648} & \textbf{0.753} & \textbf{0.624} & \textbf{0.643} \\
    \midrule
    \midrule
    \textbf{Arabic} & $\mathcal{L}$ & \multicolumn{3}{c|}{$\geq$ 1 co.emo} & \multicolumn{3}{c|}{$\geq$ 2 co.emo} & \multicolumn{3}{c}{$\geq$ 3 co.emo} \\ \midrule
    BERT$_{\text{base}}$ & BCE & 0.656 & 0.459 & 0.527 & 0.668 & 0.471 & 0.531 & 0.682 & 0.485 & 0.555 \\
    SpanEmo & BCE & \textbf{0.689} & 0.518 & \textbf{0.565} & 0.709 & 0.536 & 0.586 & 0.745 & 0.567 & \textbf{0.629} \\
    SpanEmo & Joint & \textbf{0.689} & \textbf{0.534} & \textbf{0.565} & \textbf{0.710} & \textbf{0.551} & \textbf{0.587} & \textbf{0.746} & \textbf{0.584} & 0.626 \\
    \midrule
    \midrule
    \textbf{Spanish} & $\mathcal{L}$ &  \multicolumn{3}{c|}{$\geq$ 1 co.emo} & \multicolumn{3}{c|}{$\geq$ 2 co.emo} & \multicolumn{3}{c}{$\geq$ 3 co.emo} \\ \midrule
    BERT$_{\text{base}}$ & BCE & 0.603 & 0.476 & 0.526 & 0.567 & 0.461 & 0.441 & 0.518 & 0.432 & 0.364 \\
    SpanEmo & BCE & 0.653 & 0.528 & 0.561 & 0.646 & 0.528 & 0.519 & \textbf{0.663} & 0.566 & \textbf{0.508} \\
    SpanEmo & Joint & \textbf{0.662} & \textbf{0.565} & \textbf{0.581} & \textbf{0.655} & \textbf{0.568} & \textbf{0.530} & 0.644 & \textbf{0.570} & 0.490 \\
    \bottomrule
    \end{tabular}}
    \caption{Presenting the number of co-existing emotion classes. The third row in each group corresponds to the removal of LCA loss from SpanEmo. The best results in each language group are marked in bold.}
  \label{muli-label}%
\end{table*}%

Table~\ref{sec:abldation_table} presents the results. When~\textsc{s}pan\textsc{e}mo was trained without the \textsc{lca} loss, the results dropped by 1-2\% for macro F1- and jaccard score. In addition, the results of~\textsc{s}pan\textsc{e}mo dropped by 1-2\% for two metrics apart from the macro F1-score when trained without the~\textsc{bce} loss. However, the removal of the label segment led to a much higher drop of 3-6\%.~The same patterns were also observed in the \textsc{a}rabic and \textsc{s}panish experiments. This supports our earlier hypothesis that casting the task of multi-label emotion classification as span-prediction is beneficial for improving both the representation and performance of multi-label emotion classification.

\begin{table*}[htp]
\centering
\scalebox{0.9}{
% \small
\begin{tabular}{l|l}
\hline
\textbf{Emotion} & {\textbf{Top 10 Words}} \\ \hline \hline
\textbf{anger}   & anger pissed     wrath       idiots       dammit  kicking       irritated  thrown      smashed  complain \\ \hline
\textbf{anticipation} & prediction  planning    mailsport     assumptions      upcoming     waiting     route       waited    frown    ideas     \\ \hline
\textbf{disgust} & disgusting  smashed    gross       hate         pissed  wrath         dirty      awful       vile     dumb     \\ \hline
\textbf{fear}    & nervous     fear       terror      frightening  afraid  frown         panic      terrifying  scary    dreading \\ \hline
\textbf{joy}          & happy       excitement  joyful        congratulations  glad         delightful  excited     adorable  amusing  smiling   \\ \hline
\textbf{love}    & love        sweetness  loved       hug          mate    lucky         carefree   shine       care     gracious \\ \hline
\textbf{optimism}     & optimism    integrity   salvation     persevere        perspective  bright      effort      faith     glad     lord      \\ \hline
\textbf{pessimism}    & hopeless    frown       disappointed  weary            dread        despair     depressing  chronic   suicide     pain  \\ \hline
\textbf{sadness} & sadness     frown      depressing  saddened     hurt    disappointed  weary      upset       sorrow   hate     \\ \hline
\textbf{surprise}     & stunned     awestruck   shocking      awe              mailsport    buster      genuinely   curious   hardly   believing \\ \hline
\textbf{trust}   & integrity   shine      respect     courage      sign    effort        confident  faith       easy     kindness \\ \hline
\end{tabular}}
\caption{Top 10 words associated with each corresponding emotion.}
\label{top_words}
\end{table*}

\section{Analysis}~\label{ana_sec}

\subsection{Prediction of Multiple Emotions}
We additionally validated the effectiveness of our method for learning the multiple co-existing emotions on English, Arabic and Spanish sets.~Table~\ref{muli-label} presents the results, including \textsc{bert$_{\text{base}}$}.~\textsc{s}pan\textsc{e}mo demonstrated a strong ability to handle multi-label emotion classification much better than \textsc{bert$_{\text{base}}$}. Since \textsc{bert$_{\text{base}}$} is trained only with binary cross-entropy (\textsc{bce}) loss, here we include the results of our method trained only with this loss function.~\textsc{s}pan\textsc{e}mo still achieved consistent improvement as the number of co-existing emotions increases, showing the usefulness of our method in learning multiple emotions.~Improvement can clearly be observed for~\textsc{e}nglish and~\textsc{a}rabic experiments, but not as much for~\textsc{s}panish.~This may be attributed to the high percentage of single-label data, which is around ($40\%$) for~\textsc{s}panish, while it is lower than that for both~\textsc{e}nglish and~\textsc{a}rabic.~Obviously,~\textsc{s}pan\textsc{e}mo can be used without~\textsc{lca} loss, and still obtain descent performance. Nevertheless, training our method jointly with the~\textsc{lca} loss leads to better results.

\subsection{Learning Emotion-specific Associations}

\begin{figure*}[h]
\centering
  \includegraphics[width=0.9\linewidth, trim= 0cm 0.5cm 0cm 0cm, clip]{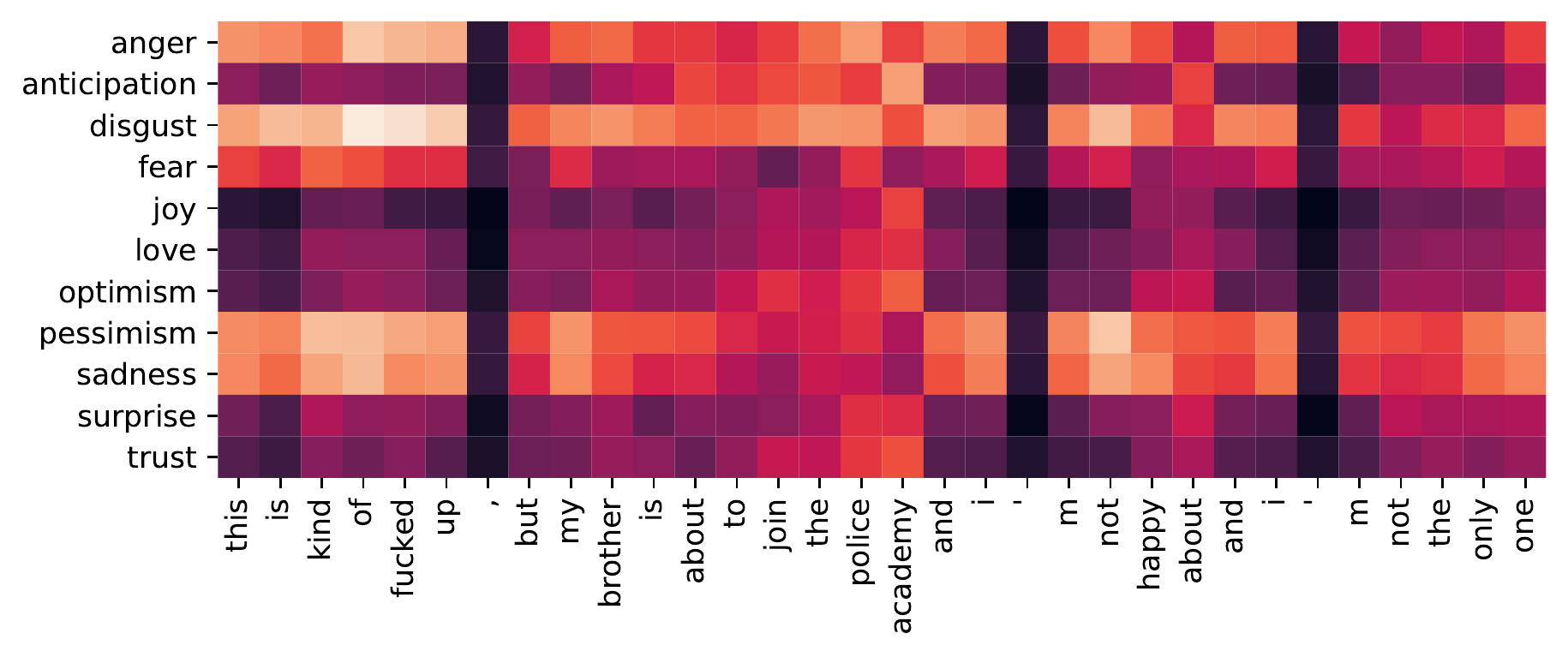}
  \caption{Visualisation on an example. The left presents the emotion labels, and the bottom presents the example. Each cell shows the cosine similarity value computed via using the hidden representation of each word and label. Lighter colour indicates higher similarity.}\label{case_study}
\end{figure*}

\subsubsection{Word-Level}\label{word_level}
%We may be able to also extract phrases for which we combine > 2 words and sum their weights ---> divide by the number of summed words (normalise version)
%We print out the top 10 key words learned from each emotion memory on SemEval2018, by first gathering the top 3 highly lighted tokens at every sen- tences on each emotion memory, and then collecting all these tokens and sorting them by their frequency. The results are shown in Table 7.

In this section, we present the top 10 words learned by \textsc{s}pan\textsc{e}mo for each emotion class by extracting the learned representations for each emotion class and all words in every input sentence, and then computing the similarity between them via cosine similarity. Finally, we performed this operation on all inputs in the \textsc{s}em\textsc{e}val2018 \textsc{e}nglish validation set and then sorted all words for each emotion class in ascending order. 

Table~\ref{top_words} presents the top-10 words per emotion class. As shown in Table~\ref{top_words}, the words discovered by our framework are indicative of the corresponding emotion.~This helps to show that \textsc{s}pan\textsc{e}mo learns meaningful associations between emotion classes and words automatically, which can be beneficial for feature extraction and learning.~Additionally, \textsc{s}pan\textsc{e}mo demonstrated that it can learn diverse words as well as shareable words across some emotions. For example, the words $\{pissed, wrath, smashed\}$ are associated with both anger and disgust, demonstrating the ability of \textsc{s}pan\textsc{e}mo to learn the relations between emotions. %We also observe that the top $30$ words include more common words across multiple emotions. 

%In this respect, we extract the learned representation from \textsc{s}pan\textsc{e}mo for each emotion class label and all words in the example. Subsequently, their similarity to each other is computed using cosine similarity. 

\subsubsection{Sentence-Level}We visualised an example from the~\textsc{e}nglish validation set annotated with four emotions, which were anger, disgust, pessimism and sadness. Our goal was to determine whether by adding emotion classes to the example, \textsc{s}pan\textsc{e}mo could learn their associations to each other (i.e., associations between emotion classes and words in the example). To compute the similarity between emotion classes and words in the example, we also followed the same process discussed in section~\ref{word_level}. 

Figure~\ref{case_study} presents the results, where lighter cells indicate higher similarity, while darker cells indicate lower similarity. As shown in Figure~\ref{case_study}, the learned representations capture the association between the correct emotion label set and every token in the example. Interestingly, we can also observe that the word ``happy'' is usually expressed as a positive emotion, but, in this context, this word becomes negative and the model learns this contextual information. Moreover, the phrase ``about to join the police academy'' is associated with ``anticipation'', which makes sense although this class is not part of the correct label set. This demonstrates the utility and advantages of our approach not only in deriving associations reported in the annotations, but also providing us with a mechanism to explore additional information beyond them.  

\subsection{Label Correlations}

\begin{figure}[!th]
\centering
\subcaptionbox{Emotion correlations: GT\label{corr:GT}}
{\includegraphics[width=\linewidth]{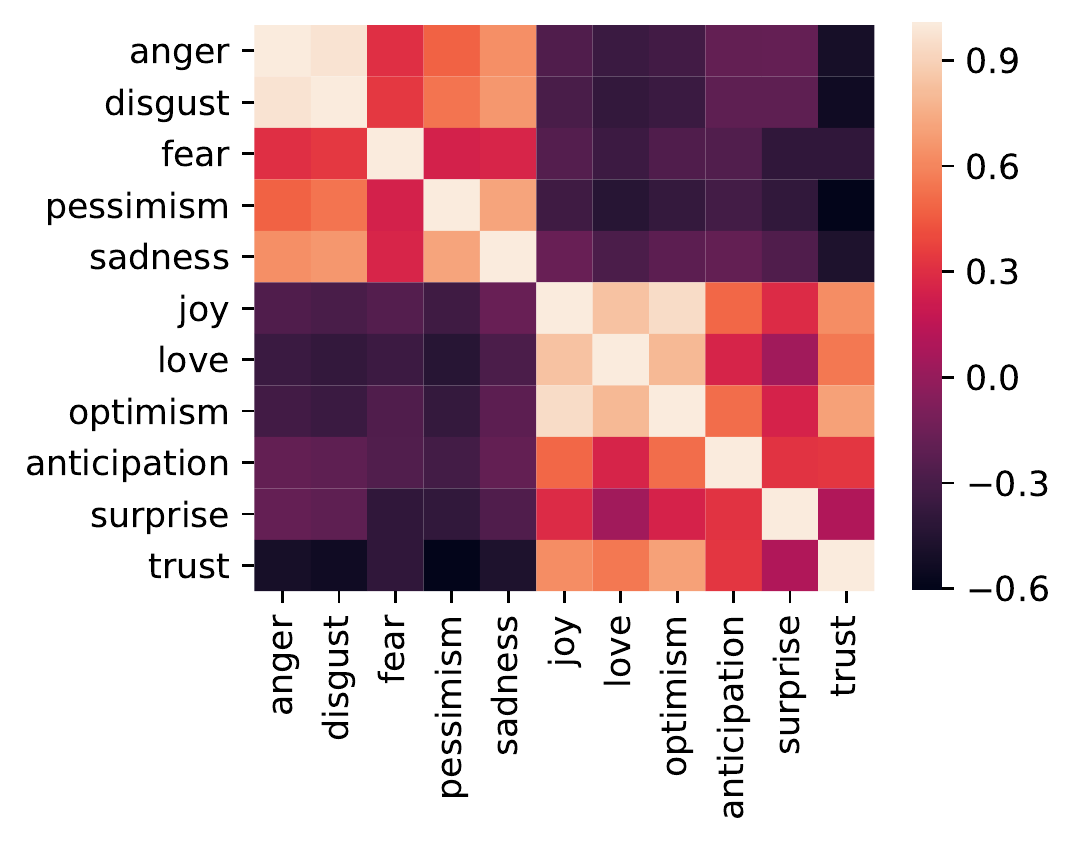}} \\
\subcaptionbox{Emotion correlations: Prediction\label{corr:pred}}
{\includegraphics[width=\linewidth]{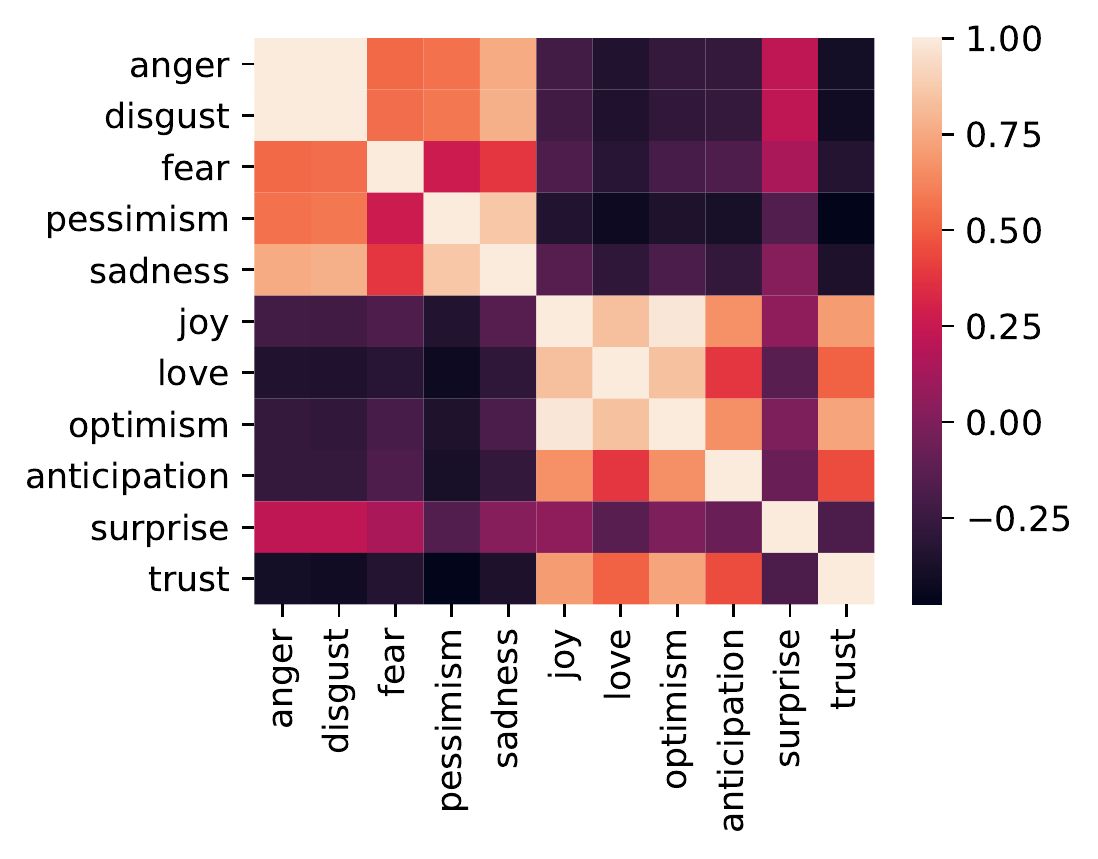}}
\caption{The top plot presents emotion correlations obtained from the ground truth (GT) labels, whereas the bottom plot presents emotion correlations obtained from the predicted labels.}\label{label_corr}
\end{figure}

Since one of the research questions in this paper was to learn the multiple co-existing emotions from a multi-label emotion data set, we analysed the learned emotion correlations from~\textsc{s}pan\textsc{e}mo and compared them to those adopted from the ground truth labels in the \textsc{s}em\textsc{e}val2018 validation set. Figure~\ref{label_corr} presents the two emotion correlations as obtained from the ground truth labels and from the predicted labels, respectively. 

% \begin{figure}[!th]
% \centering
%   \subfigure[Emotion correlations: GT]{\includegraphics[width=\linewidth]{corr_valid_y.pdf}} \\
% %   \vspace{-0.5cm}
%   \subfigure[Emotion correlations: Prediction]{\includegraphics[width=\linewidth]{corr_valid_pred_our.pdf}}
% %   \vspace{-0.4cm}
%   \caption{The top plot presents emotion correlations obtained from the ground truth (GT) labels, while the bottom plot presents emotion correlations obtained from the predicted labels.}\label{label_corr}
%   \vspace{-0.5cm}
% \end{figure}

It can be observed that Figure~\ref{label_corr}(\subref{corr:pred}) is almost identical to ~\ref{label_corr}(\subref{corr:GT}), demonstrating that our method in capturing the emotion correlations is in line with what the emotion annotations have revealed.~\ref{label_corr}(\subref{corr:pred}), which was learned by~\textsc{s}pan\textsc{e}mo, also highlights that negative emotions are positively correlated with each other, and negatively correlated with positive emotions. For example, ``anger and disgust'' share almost the same patterns, which is consistent with the studies of~\newcite{MohammadB17starsem} and~\newcite{agrawal2018learning}, both of which report the same issue with negative emotions of “anger” and “disgust”, as they are easily confused with each other. This is not surprising as their manifestation in language is quite similar in terms of the use of similar words/expression. We also noted this finding when analysing the top-10 key words learned by \textsc{s}pan\textsc{e}mo in section~\ref{word_level}. In short, taking into account emotion correlations is crucial for multi-label emotion classification in addressing the ambiguity characteristic of the task, especially for emotions that are highly correlated.  

% It can be observed that the bottom plot is almost identical to the top one, demonstrating that our method in capturing the emotion correlations is in inline with what have been revealed in the emotion annotations. The bottom plot, which is learned from~\textsc{s}pan\textsc{e}mo, also highlights that negative emotions are positively correlated with each other, while negatively correlated with positive emotions. For example, ``anger and disgust'' share almost the same patterns, which is consistent to the work of~\newcite{MohammadB17starsem} and~\newcite{agrawal2018learning} who both reported that these two emotions are highly confused with other. This is not surprising as their manifestation language is quite similar in terms of the use of similar words/expression. We also note this finding when analysing the top 10 key words learned by \textsc{s}pan\textsc{e}mo in section~\ref{word_level}. In short, taking into account emotion correlations is crucial for multi-label emotion classification in addressing the ambiguity characteristic of the task, especially for those highly correlated emotions.   

\subsection{Influence of Parameter ($\alpha$)}\label{alpha_sec}
\begin{figure}[ht]
   \includegraphics[width=\linewidth]{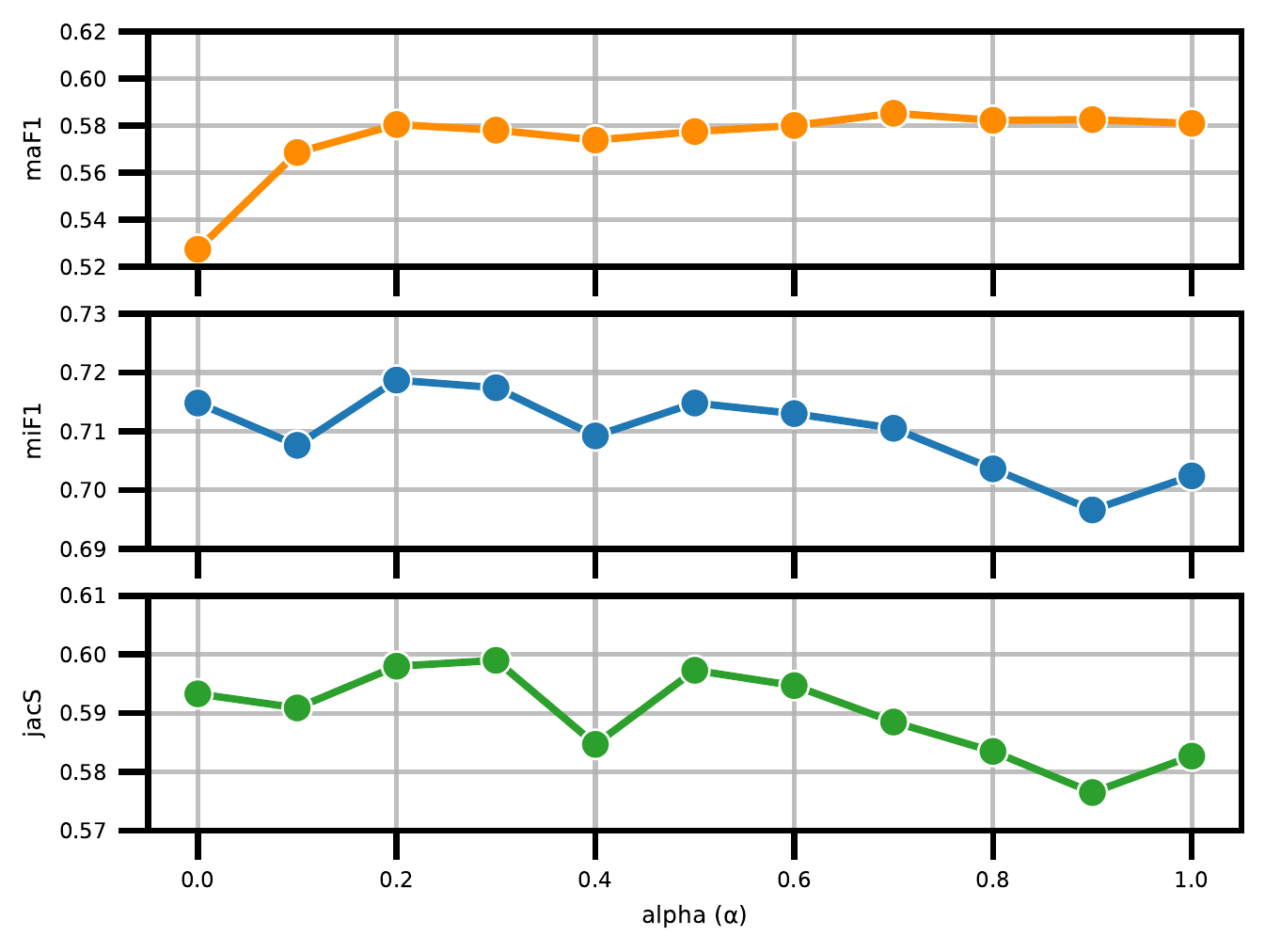}
  \caption{Sensitivity analysis of parameter ($\alpha$). Note that $\alpha = 0.0$ means that only BCE loss is used in training SpanEmo, whereas $\alpha = 1.0$ means that only LCA loss is utilised in training it.}\label{alpha}
\end{figure}

\textsc{s}pan\textsc{e}mo was trained with \textsc{bce} loss and with \textsc{lca} loss via a weight ($\alpha$), whose impact on the results is presented in Figure~\ref{alpha}. It should be mentioned that this analysis was performed on the validation set of SemEval2018 data set. The lower bound (i.e., $0.0$) indicates that the model was trained only with the \textsc{bce} loss, whereas the upper bound (i.e., $1.0$) indicates that it was trained only with the \textsc{lca} loss. When the value of $\alpha$ increased from $0.0$ to $0.5$, the results first improved considerably and then gradually deteriorated apart from the results of the macro F1-score. The results of \textsc{bce} loss favoured the micro F1- and jaccard score, whereas the results of \textsc{lca} loss favoured the macro F1-score. However, integrating \textsc{lca} with \textsc{bce} can balance the results across all three metrics, resulting in strong performance. The best results were achieved on almost all metrics when the value of $\alpha$ was set to $0.2$. Thus, we set the value of parameter $\alpha$ to $0.2$ for all experiments reported in this paper. 

\section{Related Work}\label{RW} 
There is a large body of \textsc{nlp} literature on emotion recognition~\cite{mohammad2015computational}.~Earlier studies focused on lexicon-based approaches, which make use of some words and their corresponding labels to identify emotions in text, e.g.~\textsc{nrc}\footnote{\newcite{bravo2016determining} proposes an approach for expanding it for the language used in Twitter.}~\cite{Mohammad13} and \textsc{e}mo\textsc{s}entic\textsc{n}et~\cite{poria2014emosenticspace}. Other methods treat the emotion recognition task as a supervised learning task, in which a learner (e.g. linear classifier based methods) is trained on the features of labelled data to classify inputs into one label~\cite{Bostan2018,Liew2016,mohammad2015sentiment,tang2013learning,wang2012harnessing,aman2007identifying}. %For example,~\newcite{wang2012harnessing} apply a couple of machine learning algorithms on a large \textsc{t}witter data set collected via distant-supervision by using a list of hashtags to exploit the effectiveness of the size of training data on the emotion classification. \footnote{It should be noted that the NRC lexicon have multiple emotion associations for many words.}

%~\cite{klinger-etal-2018-iest,wang2012harnessing,mohammad:2012:STARSEM-SEMEVAL,scherer1994evidence} 

%For example,~\newcite{felbo2017using} constructed a Bi-directional with self-attention mechanism on emoji’s data and then adapted it to emotion classification.~\newcite{Saravia2018} built contextualised affect representations used as features for training various recurrent neural networks and \textsc{cnn}.~\newcite{islam-etal-2019-multi} developed a multi-channel \textsc{cnn} (\textsc{m}ul-\textsc{c}h-\textsc{cnn}), which attempts to learn embeddings for each input as well as additional features occurring in the same input (e.g. emojis, emoticons and hashtags). 

More recently, several neural network models have been developed for this task, obtaining competitive results on different emotion data sets. Some of these models generally focus on a single-label emotion classification, in which only a single label is assigned to each input~\cite{islam-etal-2019-multi,xia-ding-2019-emotion,alhuzali-etal-2018-ubc,alhuzali2018enabling,agrawal2018learning,Saravia2018,felbo2017using,Abdul-Mageed2017}. Other models have also been proposed for multi-label emotion classification, in which one or more labels are assigned to each input (see detailed description in section~\ref{bl}).%~\cite{fei2020latent,baziotis2018ntua,yu2018improving,badaro-etal-2018-ema,mulki-etal-2018-tw,Mohammad2018semeval,yang2018sgm}.

Our work is motivated by research focused on learning features corresponding to each emotion as well as incorporating the relations between emotions into a loss function~\cite{fei2020latent,he2018joint}. Our work differs from these two works in the following ways: i) our method learns features related to each corresponding emotion without relying on any external resources (e.g. lexicons). ii) We further integrated the relations between emotions into the loss function by taking advantage of the label co-occurrences in a multi-label emotion data set. In this respect, our approach does not rely on any theory of emotion. iii) We empirically evaluated our method for three languages, demonstrating its effectiveness as being language agnostic. In contrast to previous research, we focus on both learning emotion-specific associations and integrating the relations between emotions into the loss function.  

\section{Conclusion} \label{conc}
We have proposed a novel framework ``\textsc{s}pan\textsc{e}mo'' aimed at casting multi-label emotion classification as a span-prediction problem.~We demonstrated that our proposed method outperforms prior approaches reported in the literature on three languages (i.e., \textsc{e}nglish, \textsc{a}rabic and \textsc{s}panish). Our empirical evaluation and analyses also demonstrated the utility and advantages of our method for multi-label emotion classification, specifically the addition of emotion classes to the input sentence, which helped the model learn emotion-specific associations and increase its performance.~Finally, training our method with \textsc{lca} loss jointly led to better results, showing the benefits of integrating the relations between emotions into the loss function. 

%including of both labels and an input sentence into the model can capture emotion-specific features as well as increase the performance

The standard approach in a multi-label emotion classification problem often focuses on modelling individual emotion independently. In this respect, most existing methods do not take into account label dependencies while learning emotion-specific associations.~However, we demonstrated the effectiveness of including label information to the input sentence when training \textsc{s}pan\textsc{e}mo, helping it achieve better performance and capture emotion correlations. We hope that this study will inspire the community to investigate further the vital role of learning label dependencies and associations corresponding to each emotion.

\section*{Acknowledgements}
We would like to thank the anonymous reviewers and Chrysoula Zerva for their valuable feedback and suggestions.~The first author is supported by a doctoral fellowship from Umm Al-Qura University.
\bibliographystyle{acl_natbib}
\bibliography{eacl2021}

\end{document}